%% file: conv_fno.tex
\documentclass[a4paper, 11pt]{article}

\usepackage{microtype}
\usepackage{graphicx}
\usepackage{subfigure}
\usepackage{booktabs} 
\usepackage{multirow}
\usepackage{makecell}
\usepackage{cite}
\usepackage{hyperref}

\usepackage{amsmath}
\usepackage{amssymb}
\usepackage{mathtools}
\usepackage{amsthm}

\usepackage[capitalize,noabbrev]{cleveref}

\theoremstyle{plain}
\newtheorem{theorem}{Theorem}[section]
\newtheorem{proposition}[theorem]{Proposition}

\theoremstyle{definition}

\theoremstyle{remark}

\newcommand{\cH}{\mathcal{H}}
\newcommand{\C}{\mathcal{C}}
\newcommand{\R}{\mathbb{R}}
\newcommand{\F}{\mathcal{F}}
\newcommand{\N}{\mathcal{N}}

\usepackage[textsize=tiny]{todonotes}

\title{Enhancing Fourier Neural Operators with Local Spatial Features}

\author{    
    Chaoyu Liu$^{1}$, Davide Murari$^{1}$, Lihao Liu$^{1}$, Yangming Li$^{1}$ \\ Chris Budd$^{2}$, Carola-Bibiane Schönlieb$^{1}$ \\
    \small $^1$Department of Applied Mathematical and Theoretical Physics, University of Cambridge \\
    \small $^2$Department of Mathematical Sciences, University of Bath \\
    \texttt{cl920@cam.ac.uk}
}

\date{}

\begin{document}

\maketitle

\input{section/0_abstract}

\input{section/1_introduction}

\input{section/2_related_work}

\input{section/3_method}


\input{section/5_experiment}

\input{section/6_conclusion}

\input{section/impact_statment}

\bibliography{reference}
\bibliographystyle{plain}

\input{section/7_appedix}

\end{document}

%% file: section/0_abstract.tex
\begin{abstract}

Partial Differential Equation (PDE) problems often exhibit strong local spatial structures, and effectively capturing these structures is critical for approximating their solutions. Recently, the Fourier Neural Operator (FNO) has emerged as an efficient approach for solving these PDE problems. By using parametrization in the frequency domain, FNOs can efficiently capture global patterns. However, this approach inherently overlooks the critical role of local spatial features, as frequency-domain parameterized convolutions primarily emphasize global interactions without encoding comprehensive localized spatial dependencies. Although several studies have attempted to address this limitation, their extracted Local Spatial Features (LSFs) remain insufficient, and computational efficiency is often compromised.
To address this limitation, we introduce a convolutional neural network (CNN)-based feature pre-extractor to capture LSFs directly from input data, resulting in a hybrid architecture termed \textit{Conv-FNO}. Furthermore, we introduce two novel resizing schemes to make our Conv-FNO resolution invariant. In this work, 
we focus on demonstrating the effectiveness of incorporating LSFs into FNOs by conducting both a theoretical analysis and extensive numerical experiments. 
Our findings show that this simple yet impactful modification enhances the representational capacity of FNOs and significantly improves performance on challenging PDE benchmarks.

\end{abstract}


%% file: section/1_introduction.tex
\section{Introduction}

Operator learning~\cite{BOULLE202483,brandstetter2022message,cao2021choose,franco2023approximation,huang2023introduction,kovachki2024operator,zhang2024biophysics} has emerged as a transformative framework for modeling complex mappings between infinite-dimensional function spaces, offering significant potential for scientific computing, engineering, and applied mathematics. Within this area, neural operators \cite{fanaskov2023spectral,kovachki2023neural,li2021fourier,lu2021learning,raonic2024convolutional,wang2021learning} represent a notable advancement in the approximation of solution operators for partial differential equations (PDEs) and other functional relationships in high-dimensional spaces.
Among these, the Fourier Neural Operator (FNO) and its variants \cite{bonev2023spherical,helwig2023group,kovachki2021universal,li2021fourier,li2021physics,liu2024domain,zou2023learning,tran2021factorized} have gained substantial attention due to its good performance on various applications, particularly for dynamic simulation and the solving of PDEs.

A key strength of FNO lies in its use of the Fourier transform \cite{cochran1967fast,kumar201950} to efficiently capture global dependencies. By parameterizing in the frequency domain, FNOs establish suitable mappings between frequency representations while bypassing the grid-dependent nature of many traditional architectures. This design grants FNOs the highly desirable property of resolution invariance \cite{kovachki2023neural,kovachki2021universal}, enabling them to generalize across varying resolutions. As a result, FNOs are particularly well-suited for applications involving heterogeneous or multiscale data. This combination of efficiency and adaptability has enabled FNOs to achieve state-of-the-art performance on numerous PDE-based benchmarks \cite{mcgreivy2024weak,takamoto2022pdebench}, positioning them as a leading solution for tackling high-dimensional problems where traditional machine learning models often struggle \cite{gopakumar2023fourier,pathak2022fourcastnet}.
Despite their success, FNOs still face a critical limitation in their design. FNOs rely on dimension lifting with point-wise operators to ensure resolution invariance, which inherently restricts their ability to extract rich Local Spatial Features (LSFs). Moreover, their dependence on frequency-domain operations further diminishes their capacity to effectively model localized structures, which are crucial for capturing the fine-grained dynamics of complex systems.

LSFs play a pivotal role in numerous scientific and engineering applications, as they encode essential fine-grained information such as sharp gradients, discontinuities, and small-scale structures. In the context of PDE solutions, LSFs are crucial for accurately capturing and governing the system's dynamics. For example, boundary layer effects in fluid dynamics, sharp phase interfaces in material science, and localized stress concentrations in structural analysis all demand accurate representation of LSFs. While the Fourier transform is highly effective at capturing global dependencies, its inherent limitations in extracting LSFs can negatively impact performance in scenarios where localized details are crucial.
Several approaches have been proposed to address this challenge, including FNO variants such as U-FNO \cite{wen2022u}. While these methods improve the extraction of local information, they often do so at the expense of the resolution-invariance property. Moreover, they primarily focus on enhancing frequency-domain feature representations without explicitly capturing localized spatial features (LSFs) from the input data. Recently, the convolutional neural operator (CNO) \cite{raonic2024convolutional} was proposed to improve LSF modeling. While effective at enhancing local spatial feature extraction, CNO sacrifices the Fourier transform, a fundamental strength of FNOs for capturing global information, thereby sacrificing one of FNO's key advantages.
More recently, the Local FNO \cite{liu-schiaffini2024neural} was introduced, incorporating differential and local integral kernels to extract local information while maintaining FNO’s resolution-invariant property. However, the differential kernels are limited to capturing only differential features, while the local integral kernels are significantly more computationally expensive compared to convolutional operators. This leaves a gap in effectively and efficiently extracting LSFs within the FNO framework. In this work, we focus on \textbf{how to effectively integrate LSFs into FNOs}. Our approach aims to enhance the ability of FNOs to balance local and global representations while preserving their computational efficiency and resolution-invariance properties.

Inspired by the success of CNO, we propose a simple yet powerful enhancement to the FNO architecture: the Convolutional Fourier Neural Operator (Conv-FNO). Specifically, our approach integrates a convolutional neural network (CNN) as a feature pre-extractor to extract LSFs directly from the input data before applying the Fourier transform. This preprocessing step effectively bridges the gap between local and global representations, allowing the model to retain fine-grained spatial details while leveraging the Fourier transform for efficient global interactions. Unlike previous methods focusing on frequency-domain modifications or downstream architectural enhancements, our approach explicitly targets input-stage feature extraction to ensure that critical local information is preserved from the outset.

Our proposed Conv-FNO preserves the resolution-invariance property of the original FNO through carefully designed resizing schemes, making it adaptable to grids of varying resolutions. Additionally, we introduce a specific variant, UNet-FNO, which incorporates a UNet-based encoder-decoder structure \cite{ronneberger2015u,liu2020psi} as the feature pre-extractor. This design leverages the hierarchical feature extraction capabilities of UNet to capture multi-scale spatial information, significantly enhancing the model’s ability to handle complex PDE solutions requiring both local and global representations.

We validate our proposed approach through extensive experiments on diverse benchmark PDE datasets, including the Allen-Cahn equation, the Navier-Stokes equation, Darcy flow, and more. The results demonstrate that Conv-FNO and its variant UNet-FNO consistently outperform existing neural operator methods, delivering superior accuracy across a wide range of scenarios. Notably, our hybrid spatial-frequency architecture demonstrates excellent performance even with limited training samples, showcasing its potential for applications where generating large training datasets is costly or resource-intensive. Our contributions are:
\begin{itemize}
\item \textbf{Efficient Integration of Comprehensive LSFs into FNOs:} We present a straightforward yet effective framework to integrate LSFs into the FNO. This addresses FNO’s inherent limitation in capturing fine-grained spatial details while preserving computational efficiency.

\item \textbf{Development of Conv-FNO:} Building upon this idea, we propose the Convolutional Fourier Neural Operator (Conv-FNO), which combines a convolutional neural network (CNN) with FNO to achieve hybrid spatial-frequency modeling. Extensive experiments on benchmark PDE problems demonstrate that Conv-FNO, particularly its UNet-based variant, consistently outperforms state-of-the-art methods.  

\item \textbf{Preservation of Resolution Invariance:} To retain the resolution-invariance property of FNO, we introduce two resizing schemes. Theoretical analysis and experimental results validate that Conv-FNO achieves robust accuracy across different grid resolutions.  

\item \textbf{Hybrid Spatial-Frequency Modeling:} Our approach bridges the gap between comprehensive local and global representations, effectively combining spatial- and frequency-domain modeling. This hybrid architecture offers superior accuracy and generalization capabilities without compromising resolution invariance.  

\item \textbf{Robustness in Data-Limited Scenarios:} The proposed Conv-FNO and UNet-FNO demonstrate exceptional performance even with limited training data, making them ideal for applications where data generation is costly or resource-intensive.  
\end{itemize}

%% file: section/2_related_work.tex
\section{Local Spatial Features}
Partial differential equations (PDEs) inherently involve a series of local operators, such as the gradient operator, divergence operator, and Laplacian operator. These operators underscore the critical role of local spatial features in PDEs. While global influences may accumulate over extended periods, the immediate evolution of a point is dominated by local interactions. In evolution equations, within a small time interval, the temporal evolution of a specific point is highly dependent on its surrounding points. Accurately capturing local spatial features not only enhances computational accuracy but also reduces the overall computational complexity. This principle is evident in numerical methods like adaptive mesh refinement, where regions of the solution with sharp interfaces or steep gradients are assigned a finer mesh, leading to improved accuracy and computational efficiency \cite{eriksson1996computational,huang2010adaptive}. 

Current neural operators often emphasize either local spatial features or global features, but rarely both. For example, Unet-based neural operators primarily focus on local spatial information, while their ability to capture global features remains limited. In contrast, the Fourier Neural Operator (FNO) processes input point-wise and subsequently transforms this information into the frequency domain using the Fast Fourier Transform (FFT). The parameterization of FNO occurs predominantly in the frequency domain, endowing it with the desirable property of capturing global features. The processor in FNO ensures resolution invaiance; however, this comes at the expense of reduced capability to capture detailed local spatial information. On the other hand, the Convolutional Neural Operator (CNO) re-emphasizes the significance of convolution operators, leveraging them to effectively capture spatial features and enhance neural operator modeling. By relying solely on spatial features, CNO demonstrates its ability to outperform FNO in certain tasks, further highlighting the critical role of local spatial features in the context of neural operators.

In this work, we propose the Conv-FNO, a novel architecture that effectively integrates LSFs into FNOs. Unlike traditional FNOs, which rely exclusively on frequency-domain parameterization and lose detailed local spatial information, the Conv-FNO introduces a convolutional neural network (CNN) as a preprocessor. This CNN module is specifically designed to extract rich LSFs directly from the input data, ensuring that fine-grained spatial details are retained from the outset. 

Moreover, the Conv-FNO architecture leverages the complementary strengths of spatial and frequency domains. While the CNN preprocessor efficiently captures the intricate local interactions, the subsequent FNO models global dependencies based on the extracted LSFs. It is worth mentioning that the valuable information from the LSFs is preserved throughout the forward process rather than being limited to the input of the FNO. This is achieved by alternating FFT and inverse FFT within the FNO, which can maintain and propagate this information.  Consequently, the hybrid design of Conv-FNO seamlessly integrates both spatial and frequency features, making it well-suited for applications that demand a balance between local precision and global contextual understanding. The following section presents a detailed overview of the Conv-FNO architecture and emphasizes its key strengths.

%% file: section/3_method.tex
\section{Conv-FNO}
\paragraph{Architecture of Conv-FNO}

The architecture of Conv-FNO can be represented by the following expression:
\begin{equation}
\label{architecture}
    \mathcal{P}_{\theta_2 }\circ \text{CAT} \circ \text{CNN}_{\theta_1}.
\end{equation}
Here, \text{CAT} is the concatenation operation concatenating the CNN input to its output along the channel dimension, $\text{CNN}_{\theta_1}$ is a convolutional neural network with trainable parameters $\theta_1$, and $\mathcal{P}_{\theta_2 }$ is a Fourier Neural Operator with trainable parameters $\theta_2$. The CNN serves to extract rich LSFs, which are subsequently concatenated with inputs along the channel dimension and fed into the FNO to model the global dynamics and produce the final outputs. For the CNN component, we experimented with simple combinations of convolutional layers with varying kernel sizes as well as a UNet-based architecture. Both approaches demonstrated significant improvements over the original FNO, with the UNet-based Conv-FNO achieving superior performance. Experimental results reveal that the Unet-FNO substantially outperforms state-of-the-art neural operators, including FNO and CNO, across a wide range of PDE benchmarks.


\subsection{Toy Conv-FNO}
To demonstrate the importance of local spatial features, we employ a minimalistic CNN architecture within Conv-FNO. This simple CNN consists of convolutional layers with different kernel sizes to capture spatial features at varying scales. The output features from all layers are first processed by two convolutional layers with kernel sizes of $\{5,3\}$, then concatenated along the channel dimension and finally fused using a $1 \times 1$ convolution to integrate multi-scale information.

\begin{figure}
\centering
\includegraphics[width=0.9\linewidth]{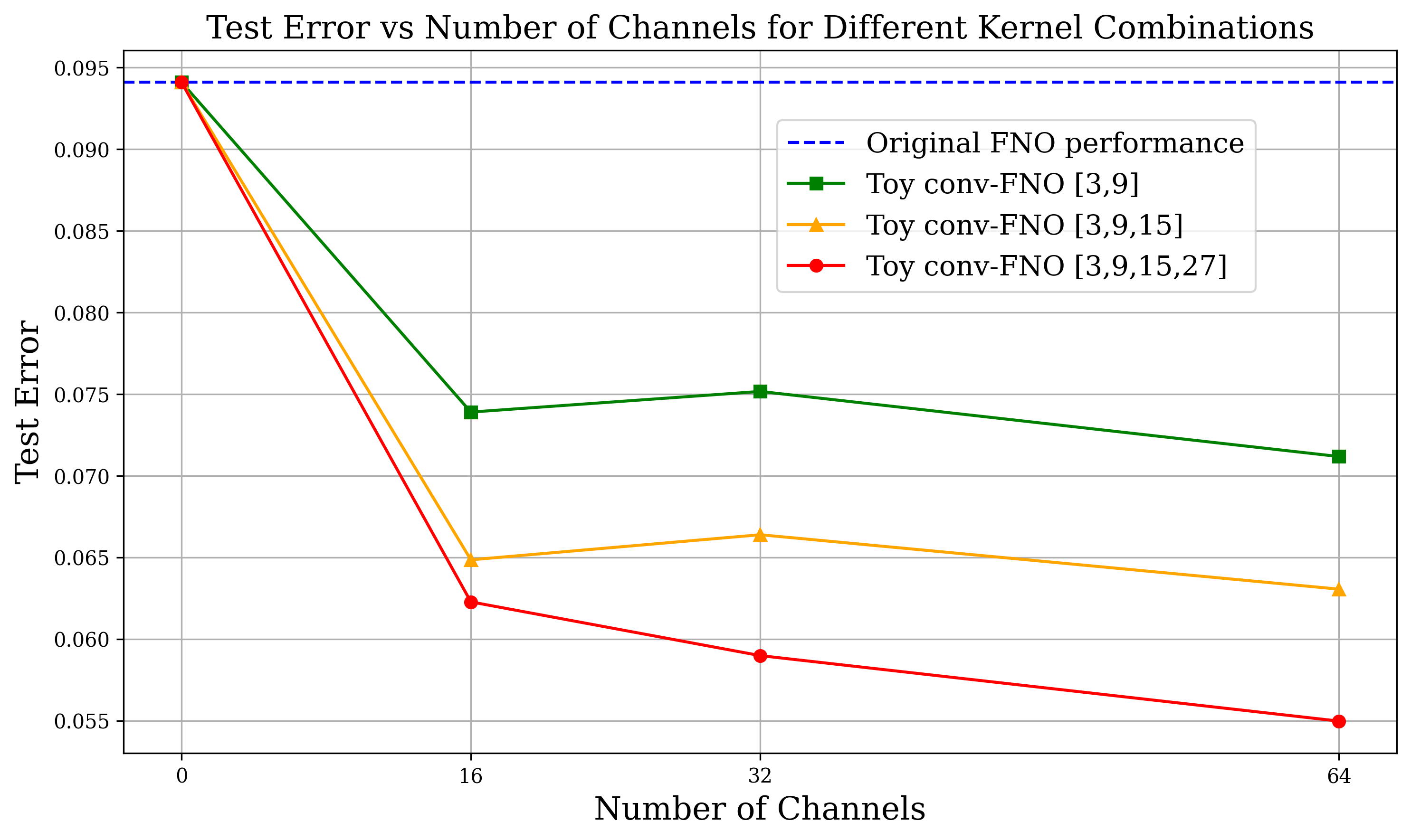}
\caption{Test error of Toy Conv-FNO with different convolutional layers and channels}
\label{fig:toy_fno}
\end{figure}

We evaluate the performance of this toy Conv-FNO on the Navier-Stokes equation ($\nu=0.0001$) and compare it against the original FNO. To analyze the impact of local spatial features (LSFs), we progressively increase the complexity of the convolutional layers, starting with kernel sizes of $\{3, 9\}$ and expanding to $\{3, 9, 15, 27\}$. Additionally, we examine the effect of varying the number of output channels in the CNN. The results in Figure \ref{fig:toy_fno} demonstrate that our toy Conv-FNO significantly outperforms the baseline FNO. Furthermore, the findings indicate that incorporating richer LSFs enhances performance. Specifically, increasing the number of LSFs and introducing multiple scales further improve FNO’s predictive accuracy. This highlights the crucial role of local spatial features in accurately modeling the dynamics of PDE systems.

\subsection{UNet-FNO}
Beyond the simple CNN, more sophisticated architectures can be integrated into Conv-FNO to capture richer and more complex LSFs. Specifically, we replace the basic CNN with the well-established UNet architecture, leading to the development of the UNet-based Conv-FNO (denoted as UNet-FNO). The specific configuration of UNet is provided in \ref{model_implementation}.

The UNet architecture enables hierarchical feature extraction through its encoder-decoder structure, allowing Conv-FNO to extract deeper and more flexible representations from the input data. Consequently, the integration of UNet leads to further improvements in predictive accuracy. Experimental results demonstrate that UNet-FNO achieves superior performance and surpasses state-of-the-art neural operators across a wide range of PDE benchmarks.

\section{Resolution Invariance}
\subsection{Resizing Schemes}
The original FNO possesses a notable property known as resolution invariance, which ensures that the prediction error remains consistent across different grid resolutions. This property arises because the Fourier layers in FNO can be regarded as discretization-invariant, and the dimension-lifting operation in FNOs is implemented via point-wise operators, which are inherently independent of grid resolution. As a result, FNOs can efficiently transfer solutions between different resolutions and perform zero-shot super-resolution.

In the case of Conv-FNOs, the introduction of standard convolutional operators (with kernel sizes larger than one) disrupts this resolution invariance because the outputs of convolution operations depend on the input grid resolution. To address this challenge and retain resolution invariance, we propose two resizing schemes for Conv-FNO. 

\paragraph{Resizing Scheme 1}
The first scheme consists of three steps. The first step is to resize the input data to match the grid resolution used during training. Conv-FNO is then applied to produce outputs at this fixed resolution. Finally, the outputs are resized back to the original resolution. The scheme 1 can be summarized by the following formula:
\begin{equation}
   R_2\circ\mathcal{P}_{\theta_2}\circ \text{CAT}\circ \text{CNN}_{\theta_1}\circ R_1.
\end{equation}
Here $R_1$ and $R_2$ are resizing operators that adjust the input resolution to the training resolution and restore the output back to the original resolution, respectively. While the downsampling step may result in the loss of fine-grained information, this method is straightforward and can be easily implemented.

\paragraph{Resizing Scheme 2}
To mitigate the limitations of Scheme 1, we propose an alternative approach that directly leverages the resolution invariance property of FNO. The scheme 2 can be described by the following expression:
\begin{equation}
\label{eq:scheme2_process}
   \mathcal{P}_{\theta_2}\circ \text{CAT}\circ R_2\circ \text{CNN}_{\theta_1}\circ R_1.
\end{equation}

As illustrated in \eqref{eq:scheme2_process}, the input is first resized to match the grid resolution used during training for the CNN to extract LSFs. The extracted LSFs are then resized back to the original resolution and concatenated with the original input before being fed into the FNO. In this scheme, only the LSFs are subject to resizing, while the original input remains unchanged when passed to the FNO. This ensures that the fine details of the input data are preserved, while still benefiting from the local feature extraction provided by the CNN. In the original FNO, the input consists of PDE data concatenated with spatial coordinates $(x, y)$. These spatial coordinates also serve as a form of spatial feature representations. When making predictions at different resolutions, these spatial features are also resized accordingly. In Conv-FNO with Scheme 2, the key distinction lies in the incorporation of additional spatial features extracted by the CNN. These extracted features, like the $(x, y)$ coordinates in FNO, are resized to the target input resolution. This ensures that Conv-FNO inherits the resolution invariance property of FNO. Experiments also show that Conv-FNO with Scheme 2 exhibits a resolution invariance behaviour that closely mirrors the one of the original FNO; see Figure \ref{fig:reoslution_invariance_test}.

Experimental results confirm that Conv-FNO achieves resolution invariance under both proposed resizing schemes, with Scheme 1 offering a straightforward approach and Scheme 2 better preserving the inherent resolution invariance of the original FNO. However, despite its theoretical advantages, Scheme 2 does not consistently outperform Scheme 1 in practice. In fact, Scheme 1 achieves slightly better performance, suggesting that the resizing in Scheme 1 does not necessarily degrade the overall accuracy.



\subsection{Theoretical Analysis}
\label{theoretical_analysis}
The goal of an FNO is to approximate an operator between two Hilbert spaces $\cH_1,\cH_2$, which we call $\C:\cH_1\to\cH_2$. We now suppose that the Hilbert space $\cH_1$ is made of functions from $\Omega\subset\R^{n}$ and taking outputs in $\R^{d_1}$. The Hilbert space $\cH_2$ takes instead inputs in ${\Omega}\subset\R^{n}$ and returns outputs in $\R^{d_2}$. This means that for an $x\in\Omega$, $u\in \cH_1$, we have $u(x)\in\R^{d_1}$ and $\C(u)(x)\in\R^{d_2}$. 

Let us now call $\chi_m$ a regular grid of $m$ nodes on $\Omega$, and call $\F_m^{c_0,h}$ the family of FNOs trained on resolution $m$ with $c_0$ input channels and $h$ output channels of the dimension lifting operator. Thus, given a function $u\in \cH_1$, we can express the restrictions $u|_{\chi_m}$ and $\C(u)|_{{\chi}_m}$ as vectors of size $md_1$ and $md_2$ respectively. This is more commonly arranged as a suitably shaped tensor, but we consider it a vector for convenience here. Let us also define $i_{m,m'}:\R^{md_1}\to \R^{m'd_1}$ and $\tilde{i}_{m,m'}:\R^{md_2}\to\R^{m'd_2}$ the up/downsampling operators for inputs and outputs, respectively. It is clear that, depending on the input $u\in\cH_1$ one considers, the reconstruction gap
\[
\|i_{m,m'}(u|_{\chi_m}) - u|_{\chi_{m'}}\|_2
\]
can vary considerably, and the same applies to $\C(u)\in\cH_2$.  To simplify the notation, let us define $U_m = u|_{\chi_m}$ and $C_m(u)=\C(u)|_{{\chi}_m}$, where $C_m:\cH_1\to\R^{d_2m}$. We also introduce the sets
\[
\mathcal{U}_m:=\left\{V\in\R^{md_1}:\,\,V=v|_{\chi_m}\text{ for a }v\in \cH_1\right\},
\]
\[
\mathcal{V}_m:=\left\{V\in\R^{md_2}:V=\mathcal{C}(v)|_{\chi_m}\text{ for a }v\in\cH_1\right\},
\]
to which $U_m$ and $C_m(u)$ belong, respectively. \newline\newline
Let us now consider an FNO $\N \in\F_m^{c_0,h}$ which, for every $u\in \mathcal{H}_1$, realizes
\begin{equation}\label{eq:FNOref}
\|\N(U_m) - C_m(u)\|_2 =: \varepsilon_m(u).
\end{equation}
By the universal approximation property of FNOs, see \cite{kovachki2021universal}, it is clear that, up to changing the considered FNO, one could make the associated error $\varepsilon_m(u)$ arbitrarily small. However, in this note, we keep it as a function and we do not focus on having necessarily small values. The FNO $\N$ can be found by training at the specific resolution determined by the grid $\chi_m$ defined over $\Omega\subset\R^{n}$.  
Suppose that when we test this FNO $\N\in\F_m^{c_0,h}$ on inputs sampled at a different grid $\chi_{m'}$ we have 
\begin{equation}\label{eq:performanceChange}
\|\N(U_{m'}) - C_{m'}(u)\|_2 =: \varepsilon_{m,m'}(u),
\end{equation}
where $\varepsilon_{m,m'}(u)>0$ will generally depend on $\varepsilon_m(u)$, on the gap between $m$ and $m'$, and on the types of functions contained in $\cH_1$. We now want to get some upper bounds on the quality of the map we can get with the two resizing schemes making Conv-FNO resolution invariant. 

\paragraph{Resizing scheme 1} We start with the resizing scheme restoring the input resolution only after the application of the FNO. In this case, the maps over which we optimise are collected in the set
\begin{equation}\label{eq:scheme1}
\begin{split}
\mathcal{R}_1 := \Big\{\tilde{i}_{m,m'} \circ \mathcal{P} \circ 
\underline{\text{CAT}} \circ \text{CNN}_{\theta} \circ i_{m',m}: \\ 
\R^{m'd_1} \to \R^{m'd_2} \,\big|\, \mathcal{P} \in \F_m^{c,h},\,\, \theta \in \R^p \Big\},
\end{split}
\end{equation}
where $c=c_0+k$, $c_0$ is the number of input channels of our data points, and $k$ is the number of output channels of the $\text{CNN}$. We remark that the $\text{CNN}$ depends on a set of trainable parameters collected in $\theta\in\R^p$, and $\underline{\text{CAT}}$ is the concatenation operation which concatenates the resized input $i_{m',m}(u|_{\chi_{m'}})\in\mathcal{U}_{m}$ with the CNN output. Furthermore, the notation $\text{CNN}$ assumes the input to be reshaped into a suitable tensor, whereas here we work with vectors for convenience.
\begin{theorem}\label{thm:scheme1}
Let $\mathcal{N}\in\mathcal{F}^{c_0,h}_m$ be an FNO realizing the errors in \eqref{eq:FNOref} and \eqref{eq:performanceChange}. The ``cross-resolution" error
\[
\begin{split}
\inf_{\text{Conv-FNO}\in\mathcal{R}_1}&\left\|\text{Conv-FNO}(u|_{\chi_{m'}}) - C_{m'}(u)\right\|_2\\
& \leq \Delta_{m,m'}(u)+\mathrm{Lip}(\tilde{i}_{m,m'})\mathrm{Lip}(\N)\delta_{m,m'}(u)\\
& + \mathrm{Lip}(\tilde{i}_{m,m'})\varepsilon_{m}(u),
\end{split}
\]
where $\mathrm{Lip}(\tilde{i}_{m,m'})$ and $\mathrm{Lip}(\N)$ are the $\ell^2-$based Lipschitz constants of $\tilde{i}_{m,m'}$ and $\N$ respectively, whereas
\[
\begin{split}
\Delta_{m,m'}(u)&:=\left\|\tilde{i}_{m,m'}(C_m(u))- C_{m'}(u)\right\|_2,\\
\delta_{m,m'}(u)&:=\left\|i_{m',m}(u|_{\chi_{m'}})-u|_{\chi_m}\right\|_2,\\
\varepsilon_m(u)&:=\left\|\N(u|_{\chi_m}) - C_m(u)\right\|_2 .
\end{split}
\]
$\Delta_{m,m'}(u)$, $\delta_{m,m'}(u)$ and $\varepsilon_m(u)$ represent the output reconstruction error, the input reconstruction error and the FNO error, respectively.
\end{theorem}
See Appendix \ref{appendix:proof} for a proof of Theorem \ref{thm:scheme1}. We remark that the Lipschitz constant of the up and downsampling layers are left generic because their magnitude depend on the discrepancy between $m$ and $m'$, on the Hilbert spaces $\mathcal{H}_1$ and $\mathcal{H}_2$, and on the technique used to define them. The analysis in Theorem \ref{thm:scheme1} suggests that the considered resizing scheme might decrease the approximation accuracy of FNOs; see \eqref{eq:FNOref}. However, we remark that our proof does not leverage the features that the CNN extracts, and this might explain the promising experimental results we still get with such a resizing scheme.
\paragraph{Resizing scheme 2} We now consider the resizing scheme where the FNO receives a concatenation of the original input with the CNN output of adapted resolution. This scheme can be described by the maps
\begin{equation}\label{eq:scheme2}
\begin{split}
\mathcal{R}_2:=\Big\{\mathcal{P} \circ \text{CAT} \circ i_{m,m'} \circ \text{CNN}_{\theta} \circ i_{m',m}: \\ 
\R^{m'd_1} \to \R^{m'd_2} \,\big|\, \mathcal{P} \in \F_m^{c,h},\,\, \theta \in \R^p \Big\},
\end{split}
\end{equation}
where we are using the same notation of \eqref{eq:scheme1}. We remark that, here, $\text{CAT}$ is the concatenation operation which concatenates the input $U_{m'}=u|_{\chi_{m'}}\in\mathcal{U}_{m'}$ with the interpolated CNN output. 
\begin{proposition}
The space of functions that the Conv-FNO in \eqref{eq:scheme2} can represent contains the one of those representable simply with the space of FNOs $\F_m^{c_0,h}$. More explicitly, $\F_m^{c_0,h}\subset\mathcal{R}_2$.
\end{proposition}
The proof is trivial, since the FNO $\mathcal{P}\in\F_m^{c,h}$ in \eqref{eq:scheme2} could be chosen to neglect the resized CNN outputs, and only consider the concatenated input, whose resolution is never changed. We can thus guarantee that for every $\N\in\F_m^{c_0,h}$ there exists an FNO $\mathcal{P}\in\F_m^{c,h}$ such that
\[
\N = \mathcal{P} \circ \text{CAT} \circ i_{m,m'} \circ \text{CNN}_\theta \circ i_{m',m},
\]
and hence the same expressivity results apply, together with the performance variations expressed by $\varepsilon_{m,m'}(u)$ in \eqref{eq:performanceChange}.

\begin{theorem}\label{thm:scheme2}
The optimal Conv-FNO can have lower error than the original FNO on all inputs across different resolutions. More explicitly,
\[ 
\begin{split}
    \inf_{\text{Conv-FNO}\in\mathcal{R}_2}\left\|\text{Conv-FNO}(u|_{\chi_{m'}}) - C_{m'}(u)\right\|_2 \\ \le \inf_{\N\in\mathcal{F}_m^{c_0,h}}\left\|\N(u|_{\chi_{m'}}) - C_{m'}(u)\right\|_2
\end{split} 
\] for any $m'$.
\end{theorem}
See Appendix \ref{appendix:proof} for a proof of Theorem \ref{thm:scheme2}.


%% file: section/5_experiment.tex
\section{Experiments}
\subsection{Model Implementation}
\label{model_implementation}
The implemented UNet model is a 4-level architecture with an initial depth of 3 encoding layers, followed by a bottleneck and 3 decoding layers. The input channels are configurable (default: 3). Starting with 16 feature channels, the number doubles at each encoding step, reaching 128 channels at the bottleneck. Downsampling is performed using max-pooling layers with a kernel size and stride of 2, while upsampling uses transposed convolutions with the same parameters. Each convolutional block consists of two convolutional layers with a kernel size of 3, circular padding, ReLU activations, and no bias. Skip connections are used to concatenate encoder features with the corresponding decoder layers, enabling the network to integrate hierarchical and spatial details effectively. The output layer uses a 1x1 convolution to adjust the feature map dimensions to the desired number. The output of the UNet is configured to have 32 channels, which are subsequently processed by an FNO consisting of 4 FNO blocks to produce the final output. The implementation of the FNO is based on the code available at \href{https://github.com/neuraloperator/neuraloperator}{https://github.com/neuraloperator/neuraloperator}. The detailed FNO configurations for the PDE benchmarks are provided in Appendix \ref{appendix:fno_configuration}.

\subsection{PDE Datasets}
We test the performance of different neural operators on a wide range of PDE datasets including the Darcy flow, the Allen-Cahn equation, the incompressible and compressible Navier-Stokes equation. All experiments are conducted on an Nvidia A100 GPU.
\paragraph{Darcy flow} The steady-state of the 2D Darcy flow equation is governed by a second-order elliptic PDE with Dirichlet boundary conditions:
\begin{equation}
\label{eq:darcy}
\begin{aligned}
        -\nabla\cdot(a(x) \nabla u(x)) &= f(x), &&x\in D \\
        u(x) &= 0, && x \in \partial D,
\end{aligned}
\end{equation}
where $u$ is the velocity field, while $a(x)$ and $f(x)$ represent the diffusion coefficient and source term, respectively. Here, the domain $D$ is chosen as $(0, 1)^2$. Our goal is to learn the mapping $\C^\dagger$ that relates the diffusion coefficient $a$ to the velocity field $u$ such that $u = \C^\dagger (a)$. In this paper, the force term $f$ is set as a constant value. For simplicity, the source term $f$ is treated as a constant. The dataset originates from \href{https://github.com/pdebench/PDEBench}{PDEbench} \cite{takamoto2022pdebench}, and the solution is obtained by evolving a time-dependent version of the equation to steady state:
\begin{equation}
    u_t -\nabla\cdot(a(x) \nabla u(x)) = f(x), \quad x\in D
\end{equation}
with random initial conditions. This temporal evolution continues until equilibrium is reached.

\paragraph{The Allen-Cahn Equation}
The 2D Allen-Cahn equation models phase separation and takes the form:
\begin{equation}
\label{eq:ac}
\begin{aligned}
    u_t &= u - u^3 + \epsilon^2\Delta u, && \mathbf{x}\in[0,1]^2, t\in(0,T)\\
    u(0,\mathbf{x}) &= u_0(\mathbf{x}), && \mathbf{x}\in[0,1]^2
\end{aligned}
\end{equation}
where $u_0\in L^2_{per}([0,1]^2, \mathbb{R})$ is the initial condition and $0 <\epsilon \ll 1$ controls the interface thickness between phases. In our experiments, $\epsilon$ is chosen to be 0.01 and 0.05, with a final time $T=16$. The operator to be learned maps $u(0, \cdot)$ to $u(0.5, \cdot)$. Data are generated using the Forward Euler method on random initial conditions with a time step of 1e-4. The dataset contains 800 time series of size $(33, 128, 128)$, split equally for training and testing.

\paragraph{The Navier-Stokes Equation} The vorticity-stream formulation of the 2D Navier-Stokes equation for incompressible fluid flow is given by:
\begin{equation}
\label{eq:ns}
    \begin{aligned}
    \omega_t &= - u\cdot\omega_x - v\cdot\omega_y + \nu\Delta\omega + f, && \mathbf{x}\in[0,1]^2, t\in(0,T)\\
    \omega &= v_x - u_y, \quad w(0,\mathbf{x}) = w_0(\mathbf{x}), && \mathbf{x}\in[0,1]^2,
\end{aligned}
\end{equation}
where $w_0\in L^2_{per}([0,1]^2, \mathbb{R})$ is the initial condition, $0< \nu \ll 1$ represents the viscosity, $(u,v)$ is the velocity field, and $f$ is the forcing term. In our experiment, we test the cases $\nu = 0.001$ and $\nu =0.0001$ with $f = 0.1(\sin(2\pi(x + y)) + \cos(2\pi(x + y)))$ and the final time $T=50$. The operator mapping to be learned is $\omega(0, \cdot)\to \omega(0.5, \cdot)$. The dataset is generated by the same numerical scheme used in the original FNO paper. More precisely, we use a pseudo-spectral method combined with the Crank-Nicolson scheme, with a time step of 1e-4. And $w_0$ is generated according to Gaussian random filed $N(0,7^{3/2}(-\Delta+49I)^{-2.5})$. We solved 200 time series of resolution $(100, 128, 128)$, with 100 for the training set and 100 for the test set, respectively. 

\paragraph{Compressible Navier-Stokes equation}
The compressible 2D Navier-Stokes equations can be described by the following PDE system,
\begin{equation}
    \label{eq:cfd}
   \begin{aligned}
       \rho_t + \nabla\cdot (\rho \mathbf{v}) &= 0, \\
       \rho(\mathbf{v}_t + \mathbf{v} \cdot \nabla\mathbf{v}) &= -\nabla p + \eta\nabla^2 \mathbf{v} + (\zeta + \eta/3)\nabla(\nabla \cdot \mathbf{v}), \\
       \partial_t \left[\epsilon +\frac{\rho v^2}{2}\right] &=
       - \nabla \cdot \left[(\epsilon + p + \frac{\rho v^2}{2})\mathbf{v} - \mathbf{v} \cdot \sigma^\prime \right],
   \end{aligned}
\end{equation}
where $\rho$ is the mass density, $\mathbf{v}$ is the velocity, $p$ is the gas pressure, $\epsilon = p/(\Gamma - 1)$ is the internal energy, $\Gamma = 5/3$, $\sigma^\prime$ is the viscous stress tensor, and $\eta, \zeta$ are the shear and bulk viscosity, respectively. For the dataset, $M, \eta, \zeta$ are chosen to be $0.1, 0.01$ and $0.01$, respectively. The dataset is directly downloaded from \href{https://github.com/pdebench/PDEBench}{PDEbench} \cite{takamoto2022pdebench}.

\subsection{Experimental Results}
\paragraph{Accuracy.} The test errors for all the selected neural operator architectures are presented in Table \ref{table:results-combined}. The results demonstrate that our proposed UNet-FNO consistently outperforms all the other architectures across all PDE benchmarks. Notably, for cases such as the Allen-Cahn equation, the UNet-FNO trained on a small dataset (1000 data pairs) achieves better performance compared to other neural operator architectures trained on a significantly larger dataset (10000 data pairs). This suggests that UNet-FNO is particularly advantageous in scenarios with limited training data, where it can achieve competitive accuracy while significantly reducing the computational cost and time associated with data generation.

Furthermore, the results indicate that the superior performance of UNet-FNO is not solely attributable to its UNet-based architecture. While the UNO also has UNet-based architecture in frequency domain, its performance exhibits a notable gap. This improvement can be primarily attributed to the effective extraction and utilization of meaningful local spatial features, which play a critical role in enhancing prediction accuracy.

\input{table/table_2}

\paragraph{Efficiency.} We evaluate the efficiency of different neural operators by recording the number of learnable parameters and the training time required for 1000 training samples on the Darcy flow equation, as shown in Table \ref{table:efficiency_ns_1000}. While FNO has significantly fewer learnable parameters compared to CNO and UNet-FNO, its training time is not substantially shorter. In contrast, despite their larger number of learnable parameters, CNO and UNet-FNO exhibit training times comparable to FNO.

\input{table/table_3}

This observation can be attributed to the computational complexity of FNO, which involves operations such as multi-layer perceptrons (MLPs) and FFTs, both of which are not linear in computational cost. On the other hand, CNO and UNet-FNO primarily rely on convolutional operations, which are generally more computationally efficient.

The increase in training time for UNet-FNO compared to FNO is minimal, as the UNet is computationally efficient and the UNet component in the UNet-FNO is designed to be lightweight. Furthermore, we examine the impact of increasing the number of layers or the dimension-lifting channels in the original FNO, with the results summarized in Table \ref{table:fno-layers}. While these modifications significantly increase the training time and model complexity, they do not lead to a notable improvement in performance, whereas UNet-FNO consistently outperforms the original FNO.

\input{table/table_4}

\begin{figure}
\centering
\includegraphics[width=0.9\linewidth]{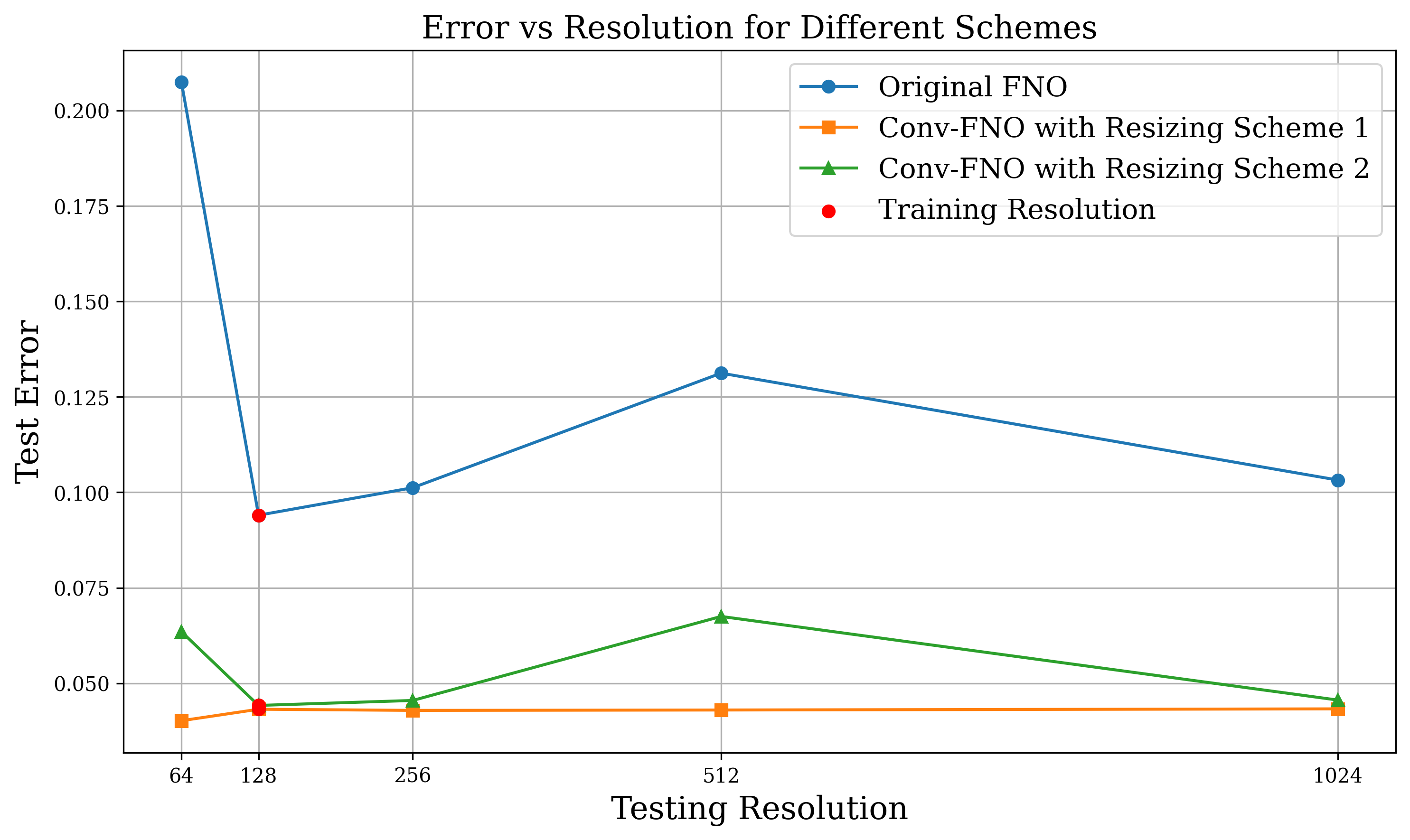}
\caption{ Test error vs. Resolution for FNO and UNet-FNO with differnt resizing schemes.}
\label{fig:reoslution_invariance_test}
\end{figure}

\paragraph{Resolution Invariance.} We evaluate the resolution invariance of both FNO and UNet-FNO on the Navier-Stokes equation ($\nu=0.0001$). The results, presented in Figure \ref{fig:reoslution_invariance_test}, demonstrate that UNet-FNO, when paired with either resizing Scheme 1 or Scheme 2, maintains strong predictive performance across test datasets with varying resolutions. Notably, both resizing schemes effectively adapt the model to inputs of different resolutions while maintaining high predictive accuracy. Scheme 1 follows a straightforward approach, resizing the input to match the training resolution before applying the model and then upsampling the output back to the original resolution. While this process may lead to some loss of fine-grained information due to downsampling, it still yields consistent results. In contrast, Scheme 2 retains the original resolution in its input and leverages the inherent resolution invariance of FNO, which further preserves fine-grained details and ensures satisfactory performance across different grid resolutions. These findings highlight that UNet-FNO, equipped with either resizing scheme, can generalize effectively to datasets with unseen resolutions, demonstrating its effectiveness in solving PDEs on variable grids.

\paragraph{Local FNO vs. Conv-FNO.} We evaluate the performance of local FNO and Conv-FNO on both accuracy and computational efficiency. The results, presented in Table \ref{tab:localfno_vs_convfno_accuracy} and Table \ref{tab:localfno_vs_convfno_efficiency}, demonstrate that UNet-FNO generally achieves better accuracy while maintaining lower computational costs. The improved performance and reduced training time highlight the advantages of incorporating convolutional preprocessing for efficiently extracting comprehensive LSFs.

Furthermore, our proposed Conv-FNO framework can be seamlessly applied to Local FNO, resulting in a hybrid model called UNet-LocalFNO. As shown in the results, UNet-LocalFNO consistently surpasses the original Local FNO in performance across various benchmark tasks. This suggests that integrating convolutional preprocessor effectively enhances Local FNO’s ability to capture more comprehensive spatial details. These findings emphasize the versatility and effectiveness of Conv-FNO in improving both accuracy and efficiency across various PDE tasks.

\input{table/table_6}
\input{table/table_7}

%% file: table/table_2.tex
\begin{table*}[htbp]
\centering
\caption{Test error of neural operators on PDE datasets with 1000 and 10000 training samples.}
\vskip 0.15in
\label{table:results-combined}
\resizebox{0.8\textwidth}{!}{
\begin{sc}
\begin{tabular}{llcccc}
\toprule
\textbf{PDE Dataset}            & \textbf{Training Samples} & \textbf{FNO} & \textbf{UNO} & \textbf{CNO} & \textbf{UNet-FNO} \\ \midrule
\multirow{2}{*}{Darcy Flow}      & 1000                     & 3.157e-2     & 3.141e-2     & 2.295e-2     & \textbf{1.312e-2} \\ 
                                 & 5000                    & 1.686e-2     & 1.770e-2     & 1.034e-2     & \textbf{7.142e-3} \\ \midrule
\multirow{2}{*}{Navier-Stokes ($\nu =1e-3$)} & 1000                     & 7.940e-3     & 7.775e-3     & 1.090e-2     & \textbf{4.250e-3} \\ 
                                 & 10000                    & 2.626e-3     & 1.937e-3     & 2.016e-3     & \textbf{1.204e-3} \\ \midrule
\multirow{2}{*}{Navier-Stokes ($\nu =1e-4$)} & 1000                     & 9.410e-2     & 9.282e-2     & 6.701e-2     & \textbf{4.135e-2} \\ 
                                 & 10000                    & 5.271e-2      & 5.617e-2     & 2.036e-2     & \textbf{1.570e-2} \\ \midrule
\multirow{2}{*}{Allen-Cahn ($\epsilon=0.05$)} & 1000                     & 1.376e-2     & 1.646e-2     & 2.980e-2     & \textbf{5.008e-3} \\ 
                                 & 10000                    & 2.945e-3     & 2.967e-3     & 1.321e-2     & \textbf{2.060e-3} \\ \midrule
\multirow{2}{*}{Allen-Cahn ($\epsilon=0.01$)} & 1000                     & 1.050e-2     & 1.511e-2     & 1.910e-2     & \textbf{3.347e-3} \\ 
                                 & 10000                    & 7.098e-3     & 1.173e-2     & 9.981e-3     & \textbf{1.135e-3} \\ \midrule
\multirow{2}{*}{Compressible Navier-Stokes} & 1000                     & 2.740e-1     & 2.846e-1     & 5.644e-1     & \textbf{2.355e-1} \\ 
                                 & 10000                    & 2.330e-1     & 2.089e-1     & 2.911e-1     & \textbf{1.640e-1} \\ \bottomrule
\end{tabular}
\end{sc}
}
\end{table*}

%% file: table/table_3.tex
\begin{table}[htbp]
\caption{Parameters and training time of different neural operators for Darcy flow dataset.}
\label{table:efficiency_ns_1000}
\begin{center}
\resizebox{0.88\columnwidth}{!}{
\begin{sc}
\begin{tabular}{lllll}
\toprule
 & FNO &UNO & CNO & Unet-FNO \\
\midrule
        Parameters & 56399 &73428 & 2951249 & 540049  \\
        Training time & 1207s &1251s & 1232s & 1492s  \\
\bottomrule
\end{tabular}
\end{sc}
}
\end{center}
\end{table}

%% file: table/table_4.tex
\begin{table}[htbp]
\caption{The comparison of FNO with different configurations and UNet-FNO on Navier-Stokes equation ($\nu=0.0001$, 1000 training samples) in terms of parameters, training time, and accuracy. The FNO in UNet-FNO has 256 lifting channels and 4 Fourier blocks.}
\label{table:fno-layers}
\begin{center}
\resizebox{1\columnwidth}{!}{
\begin{sc}
\begin{tabular}{llllll}
\toprule
 Configuration & \makecell{256-lift} &\makecell{512-lift} & \makecell{6-block} & \makecell{8-block} & \makecell{Unet-FNO} \\
\midrule
        Parameters & 73807 &242747 & 92611 & 111371 &561169  \\
        Training time & 1156s &2156s &1581s & 2017s & 1409s  \\
        Test error &0.0941 &0.0861 &0.0692 &0.0547 &\textbf{0.0414} \\
\bottomrule
\end{tabular}
\end{sc}
}
\end{center}
\end{table}

%% file: table/table_6.tex
\begin{table}
    \caption{Test error of the LocalFNO and Conv-FNOs on different PDE datasets.}
    \label{tab:localfno_vs_convfno_accuracy}
    \centering
    \resizebox{1\columnwidth}{!}{
    \begin{sc}
    \begin{tabular}{lcccc}
        \toprule
        & FNO & LocalFNO & UNet-FNO & UNet-LocalFNO \\
        \midrule
        Darcy Flow     & 1.686E-2& 9.029E-3 &7.142E-3 &\textbf{5.699E-3} \\
        Allen-Cahn $(\epsilon = 0.05)$    & 2.945E-3& 1.834E-3& 2.060E-3& \textbf{9.397E-4} \\
        Navier-Stokes $(\nu = 1e - 4)$ & 5.271E-2& 2.301E-2& \textbf{1.570E-2}& 1.636E-2 \\
        \bottomrule
    \end{tabular}
    \end{sc}}
\end{table}

%% file: table/table_7.tex
\begin{table}[htbp]
    \caption{Training time of the LocalFNO and Conv-FNOs for Darcy flow dataset (1000 samples).}
    \label{tab:localfno_vs_convfno_efficiency}
    \begin{center}
    \resizebox{0.88\columnwidth}{!}{
    \begin{sc}
    \begin{tabular}{lllll}
    \toprule
    & FNO  & LocalFNO & UNet-FNO & UNet-LocalFNO \\
    \midrule
            Training time & 1207s  &2163s & 1492s & 2524s   \\
    \bottomrule
    \end{tabular}
    \end{sc}
    }
    \end{center}
    \end{table}

%% file: section/6_conclusion.tex
\section{Conclusion}

In this work, we introduced an enhancement to the Fourier Neural Operator (FNO) by incorporating Local Spatial Features through a convolutional neural network preprocessor, resulting in the Conv-FNO architecture. Our method addresses the inherent limitation of FNOs in capturing comprehensive local details, while preserving their resolution invariance. Moreover, to maintain resolution invariance, we also proposed two novel resizing schemes. Theoretical analysis and extensive numerical experiments on benchmark PDE problems demonstrate that Conv-FNO consistently outperforms existing methods, offering superior accuracy and efficiency. By seamlessly combining spatial and frequency-domain modeling, our approach provides an effective solution for tackling complex PDEs, especially in scenarios involving limited data or multiscale structures.

%% file: section/impact_statment.tex
\section*{Impact Statement}
This paper presents work that aims to accelerate and improve the performance of Neural Operator methods. Such methods are important in the solution of partial differential equations.
The potential applications and benefits of advancements in this field are thus significant as such complex nonlinear PDEs appear everywhere in nature.

%% file: section/7_appedix.tex
\newpage
\appendix
\onecolumn

\section{Proof for the theorems in \ref{theoretical_analysis}.}
\label{appendix:proof}
\begin{theorem}
Let $\mathcal{N}\in\mathcal{F}_m^{c_0,h}$ be an FNO realizing the bounds in \eqref{eq:FNOref} and \eqref{eq:performanceChange}. The ``cross-resolution” error
\[
\begin{split}
\inf_{\text{Conv-FNO}\in\mathcal{R}_1}&\left\|\text{Conv-FNO}(u|_{\chi_{m'}}) - C_{m'}(u)\right\|_2\\
& \leq \Delta_{m,m'}(u)+\mathrm{Lip}(\tilde{i}_{m,m'})\mathrm{Lip}(\N)\delta_{m,m'}(u) + \mathrm{Lip}(\tilde{i}_{m,m'})\varepsilon_{m}(u),
\end{split}
\]
where $\mathrm{Lip}(\tilde{i}_{m,m'})$ and $\mathrm{Lip}(\N)$ are the $\ell^2-$based Lipschitz constants of $\tilde{i}_{m,m'}$ and $\N$ respectively, whereas
\[
\begin{split}
\Delta_{m,m'}(u)&:=\left\|\tilde{i}_{m,m'}(C_m(u))- C_{m'}(u)\right\|_2 \text{ is the output reconstruction error},\\
\delta_{m,m'}(u)&:=\left\|i_{m',m}(u|_{\chi_{m'}})-u|_{\chi_m}\right\|_2 \text{ is the input reconstruction error},\\
\varepsilon_m(u)&:=\left\|\N(u|_{\chi_m}) - C_m(u)\right\|_2 \text{ is the FNO error}.
\end{split}
\]
\end{theorem}

\begin{proof}
Thanks to the $\underline{\text{CAT}}$ operation, the CNN outputs could be neglected in the sense that there is an FNO $\mathcal{P}\in\F_{m}^{c,h}$ defining a Conv-FNO of the form $\tilde{i}_{m,m'}\circ \mathcal{P}\circ \underline{\text{CAT}}\circ \text{CNN}_{\theta}\circ i_{m',m} = \tilde{i}_{m,m'}\circ \N \circ i_{m',m}$.

Let us now consider the “cross-resolution” error
\[
\begin{split}
&\inf_{\text{Conv-FNO}\in\mathcal{R}_1}\left\|\text{Conv-FNO}(u|_{\chi_{m'}}) - C_{m'}(u)\right\|_2\leq\left\|\tilde{i}_{m,m'}\circ \N \circ i_{m',m}(u|_{\chi_{m'}}) - C_{m'}(u)\right\|_2 \\
&= \left\|\tilde{i}_{m,m'}\circ \N \circ i_{m',m}(u|_{\chi_{m'}})  - \tilde{i}_{m,m'}(C_m(u))+ \tilde{i}_{m,m'}(C_m(u))- C_{m'}(u)\right\|_2\\
&\leq \left\|\tilde{i}_{m,m'}\circ \N \circ i_{m',m}(u|_{\chi_{m'}})  - \tilde{i}_{m,m'}(C_m(u))\right\|_2 + {\left\|\tilde{i}_{m,m'}(C_m(u))- C_{m'}(u)\right\|_2}\\
&=\Delta_{m,m'}(u) +  \left\|\tilde{i}_{m,m'}\circ \N \circ i_{m',m}(u|_{\chi_{m'}})  - \tilde{i}_{m,m'}(C_m(u))\right\|_2 \\
&\leq \Delta_{m,m'}(u) + \mathrm{Lip}(\tilde{i}_{m,m'})\left\|\N \circ i_{m',m}(u|_{\chi_{m'}}) - C_m(u)\right\|_2\\
&=\Delta_{m,m'}(u) + \mathrm{Lip}(\tilde{i}_{m,m'})\left\|\N \circ i_{m',m}(u|_{\chi_{m'}}) - \N(u|_{\chi_m}) + \N(u|_{\chi_m}) - C_m(u)\right\|_2\\
&\leq \Delta_{m,m'}(u) + \mathrm{Lip}(\tilde{i}_{m,m'})\left(\mathrm{Lip}(\N){\left\|i_{m',m}(u|_{\chi_{m'}})-u|_{\chi_m}\right\|_2} +{\left\|\N(u|_{\chi_m}) - C_m(u)\right\|_2}\right)\\
&=\Delta_{m,m'}(u)+\mathrm{Lip}(\tilde{i}_{m,m'})\mathrm{Lip}(\N)\delta_{m,m'}(u) + \mathrm{Lip}(\tilde{i}_{m,m'})\varepsilon_{m}(u).
\end{split}
\]
\end{proof}

\begin{theorem}
The optimal Conv-FNO obtainable with the second resizing scheme can have lower error than the original FNO on all inputs across different resolutions. More explicitly,
\[ 
\begin{split}
    \inf_{\text{Conv-FNO}\in\mathcal{R}_2}\left\|\text{Conv-FNO}(u|_{\chi_{m'}}) - C_{m'}(u)\right\|_2  \le \inf_{\N\in\mathcal{F}_m^{c_0,h}}\left\|\N_m(u|_{\chi_{m'}}) - C_{m'}(u)\right\|_2
\end{split} 
\] for any $m'$.
\end{theorem}

\begin{proof}
\[
\begin{split}
        \inf_{\text{Conv-FNO}\in\mathcal{R}_2}\left\|\text{Conv-FNO}(u|_{\chi_{m'}}) - C_{m'}(u)\right\|_2 
        &= \inf_{\theta \in \R^p}\inf_{\mathcal{P}\in\mathcal{F}_m^{c,h}}\left\|\mathcal{P} \circ \text{CAT} \circ i_{m,m'} \circ \text{CNN}_{\theta} \circ i_{m',m}(u|_{\chi_{m'}})- C_{m'}(u)\right\|_2\\
        &\le \inf_{\mathcal{P}\in\mathcal{F}_m^{c,h}}\left\|\mathcal{P}\circ \text{CAT}\circ i_{m,m'} \circ  \text{CNN}_{\hat{\theta}} \circ i_{m',m} (u|_{\chi_{m'}})- C_{m'}(u)\right\|_2\\
        &= \inf_{\N\in\mathcal{F}_m^{c_0,h}}\left\|\N(u|_{\chi_{m'}}) - C_{m'}(u)\right\|_2,
\end{split}
\]
where $\text{CNN}_{\hat{\theta}}$ is returning all zeros.
\end{proof}

\section{FNO configurations for different PDE benchmarks.}
\label{appendix:fno_configuration}
    \begin{table}[h]
    \centering
    \caption{FNO Model Hyperparameters for Darcy Flow}
    \label{tab:fno_parameters}
    \begin{tabular}{l l}
        \toprule
        \textbf{Parameter} & \textbf{Value} \\
        \midrule
        n\_modes\_height       & 32 \\
        n\_modes\_width        & 32 \\
        in\_channels          & 3 \\
        lifting\_channels     & 128 \\
        hidden\_channels      & 64 \\
        out\_channels         & 1 \\
        projection\_channels  & 128 \\
        n\_layers            & 4 \\
        norm                 & group\_norm \\
        skip                 & linear \\
        use\_mlp             & true \\
        factorization        & Tucker \\
        rank                 & 1 \\
        \bottomrule
    \end{tabular}
\end{table}

    \begin{table}[h]
    \centering
    \caption{FNO Model Hyperparameters for Allen-Cahn Equation}
    \label{tab:fno_parameters}
    \begin{tabular}{l l}
        \toprule
        \textbf{Parameter} & \textbf{Value} \\
        \midrule
        n\_modes\_height       & 48 \\
        n\_modes\_width        & 48 \\
        in\_channels          & 3 \\
        lifting\_channels     & 256 \\
        hidden\_channels      & 64 \\
        out\_channels         & 1 \\
        projection\_channels  & 128 \\
        n\_layers            & 4 \\
        norm                 & group\_norm \\
        skip                 & linear \\
        use\_mlp             & true \\
        factorization        & Tucker \\
        rank                 & 1 \\
        \bottomrule
    \end{tabular}
\end{table}

    \begin{table}[h]
    \centering
    \caption{FNO Model Hyperparameters for Navier-Stokes Equation}
    \label{tab:fno_parameters}
    \begin{tabular}{l l}
        \toprule
        \textbf{Parameter} & \textbf{Value} \\
        \midrule
        n\_modes\_height       & 48 \\
        n\_modes\_width        & 48 \\
        in\_channels          & 3 \\
        lifting\_channels     & 256 \\
        hidden\_channels      & 64 \\
        out\_channels         & 1 \\
        projection\_channels  & 256 \\
        n\_layers            & 4 \\
        norm                 & group\_norm \\
        skip                 & linear \\
        use\_mlp             & true \\
        factorization        & Tucker \\
        rank                 & 1 \\
        \bottomrule
    \end{tabular}
\end{table}

    \begin{table}[h]
    \centering
    \caption{FNO Model Hyperparameters for Compressible Navier-Stokes Equation}
    \label{tab:fno_parameters}
    \begin{tabular}{l l}
        \toprule
        \textbf{Parameter} & \textbf{Value} \\
        \midrule
        n\_modes\_height       & 48 \\
        n\_modes\_width        & 48 \\
        in\_channels          & 6 \\
        lifting\_channels     & 256 \\
        hidden\_channels      & 64 \\
        out\_channels         & 1 \\
        projection\_channels  & 256 \\
        n\_layers            & 4 \\
        norm                 & group\_norm \\
        skip                 & linear \\
        use\_mlp             & true \\
        factorization        & Tucker \\
        rank                 & 1 \\
        \bottomrule
    \end{tabular}
\end{table}